%% file: main.tex
\documentclass[letterpaper, 10 pt, conference]{src/ieeeconf}  

\IEEEoverridecommandlockouts                              

\usepackage{graphics} 
\usepackage{epsfig} 
\usepackage{times} 
\usepackage{amsmath} 
\usepackage{amssymb}  
\usepackage{colortbl}
\usepackage[hidelinks]{hyperref}
\usepackage{usebib}
\usepackage[capitalise]{cleveref}
\usepackage{colortbl}
\usepackage{siunitx}
\DeclareSIUnit\mph{mph}
\usepackage{dblfloatfix}
\usepackage{soul}

\usepackage{makecell}
\usepackage{booktabs}

\usepackage{todonotes}

\newcommand{\citet}[1]{\citeauthor{#1} \cite{#1}}

\title{\LARGE \bf
SCOUT: A Lightweight Framework for \\ Scenario Coverage Assessment in Autonomous Driving}

\author{Anil Yildiz$^{*,\dagger}$, Sarah M. Thornton$^{\dagger}$, Carl Hildebrandt$^{\dagger}$, Sreeja Roy-Singh$^{\dagger}$ and Mykel J. Kochenderfer$^{*}$
\thanks{$^{*}$Department of Aeronautics and Astronautics, Stanford University,\linebreak 496 Lomita Mall, Stanford, CA 94305, \{yildiz, mykel\}@stanford.edu}%
\thanks{$^{\dagger}$Nuro Inc., 1300 Terra Bella Ave, Mountain View, CA 94043,\linebreak \{ayildiz, sthornton, childebrandt, sreeja\}@nuro.ai}%
}

\usepackage{bbm}
\usepackage[capitalise]{cleveref}
\usepackage{graphics} 
\usepackage{epsfig} 
\usepackage{amsmath} 
\usepackage{subfigure}

\usepackage[backend=bibtex,style=ieee,mincitenames=1,maxcitenames=2,maxbibnames=20,url=false,doi=false]{biblatex}
\usepackage{flushend}

\begin{document}

\maketitle
\thispagestyle{empty}
\pagestyle{empty}

\begin{abstract}
Assessing scenario coverage is crucial for evaluating the robustness of autonomous agents, yet existing methods rely on expensive human annotations or computationally intensive Large Vision-Language Models (LVLMs). These approaches are impractical for large-scale deployment due to cost and efficiency constraints. To address these shortcomings, we propose \textbf{SCOUT} ({Scenario Coverage Oversight and Understanding Tool}), a lightweight surrogate model designed to predict scenario coverage labels directly from an agent’s latent sensor representations. 
SCOUT is trained through a distillation process, learning to approximate LVLM-generated coverage labels while eliminating the need for continuous LVLM inference or human annotation. By leveraging precomputed perception features, SCOUT avoids redundant computations and enables fast, scalable scenario coverage estimation. 
We evaluate our method across a large dataset of real-life autonomous navigation scenarios, demonstrating that it maintains high accuracy while significantly reducing computational cost. Our results show that SCOUT provides an effective and practical alternative for large-scale coverage analysis. While its performance depends on the quality of LVLM-generated training labels, SCOUT represents a major step toward efficient scenario coverage oversight in autonomous systems.
\end{abstract}

\input{Sections/Ch1_Introduction}

\input{Sections/Ch2_RelatedWork}

\input{Sections/Ch3_Coverage}
\input{Sections/Ch4_LVLMs}

\input{Sections/Ch5_Method}
\input{Sections/Ch6_Experiments}
\input{Sections/Ch7_Conclusion}

\section*{Acknowledgments}
The research reported in this work was supported by Nuro~Inc. through the real-world driving scenes provided.\linebreak
We also thank Daniel Bonny, Joel Bernier, Kyle Foss, Al Ricciardelli and Ananth Kini for their valueable insights and contributions.

\renewcommand*{\bibfont}{\small}
\printbibliography
\end{document}

%% file: Sections/Ch1_Introduction.tex
\section{Introduction}

Ensuring comprehensive \textit{scenario coverage} is a fundamental challenge in evaluating the robustness and reliability of autonomous agents. Coverage analysis determines whether an agent has encountered a sufficient diversity of critical situations, particularly those involving potential failure modes or rare edge cases. However, existing methods for scenario coverage assessment typically rely on human-annotated data or high-fidelity simulation environments~\cite{zhong2021survey}, both of which are expensive and infeasible to scale. The lack of a scalable, lightweight approach to coverage estimation hinders progress in evaluating empirical safety metrics for high-risk autonomous robotics applications~\cite{ choset2001coverage}.

This problem is particularly important in safety-critical applications such as autonomous driving, robotics, and embodied AI, where failures in underexplored scenarios can lead to catastrophic consequences~\cite{khalastchi2018fault}. Without an efficient way to oversee coverage distribution, agents risk encountering novel, high-risk situations in real-world deployments without sufficient prior exposure during training or validation. A scalable coverage estimation framework would enable proactive failure mitigation, targeted policy refinements, and improved robustness across a broader set of deployment conditions. Additionally, effective scenario coverage estimation is crucial for both public trust and regulatory compliance, as agencies increasingly require systematic validation of autonomous systems before deployment~\cite{fisher2021towards}.

Traditional methods for scenario coverage estimation are computationally demanding and require heuristics and extensive human intervention~\cite{ding2023survey}. Supervised learning approaches depend on costly human-annotated datasets, which are limited in scope and difficult to scale. Large Vision-Language Models (LVLMs)~\cite{qi2023loggpt,tian2024supervised} have emerged as a potential solution, offering automated labeling capabilities, but they are prohibitively expensive to run at large scales. Furthermore, direct inference from raw sensor observations is computationally inefficient and redundant, as most perception stacks already compute feature-space representations for downstream tasks. A naive approach relying solely on human or LVLM-based annotation is therefore impractical due to cost, scalability, and efficiency constraints.

Existing solutions fail to fully address the challenges of scalable and accurate scenario coverage estimation, including the high cost of human annotation, the computational expense of LVLM-based labeling, and the inefficiencies of processing raw sensor observations. While LVLMs can augment coverage labels, their computational cost makes them unsuitable for real-time or large-scale use. On the other hand, handcrafted metrics and heuristic-based methods lack the generalizability and adaptability required for diverse and evolving deployment environments. Prior work has not effectively leveraged the precomputed latent representations from perception pipelines, missing an opportunity to efficiently infer coverage without redundant computation. To overcome these shortcomings, a lightweight, scalable alternative is needed, one that avoids direct reliance on human annotation or LVLM inference while still maintaining high coverage labeling accuracy.

\begin{figure*}[t!]
    \centering
    \includegraphics[width=0.89\linewidth]{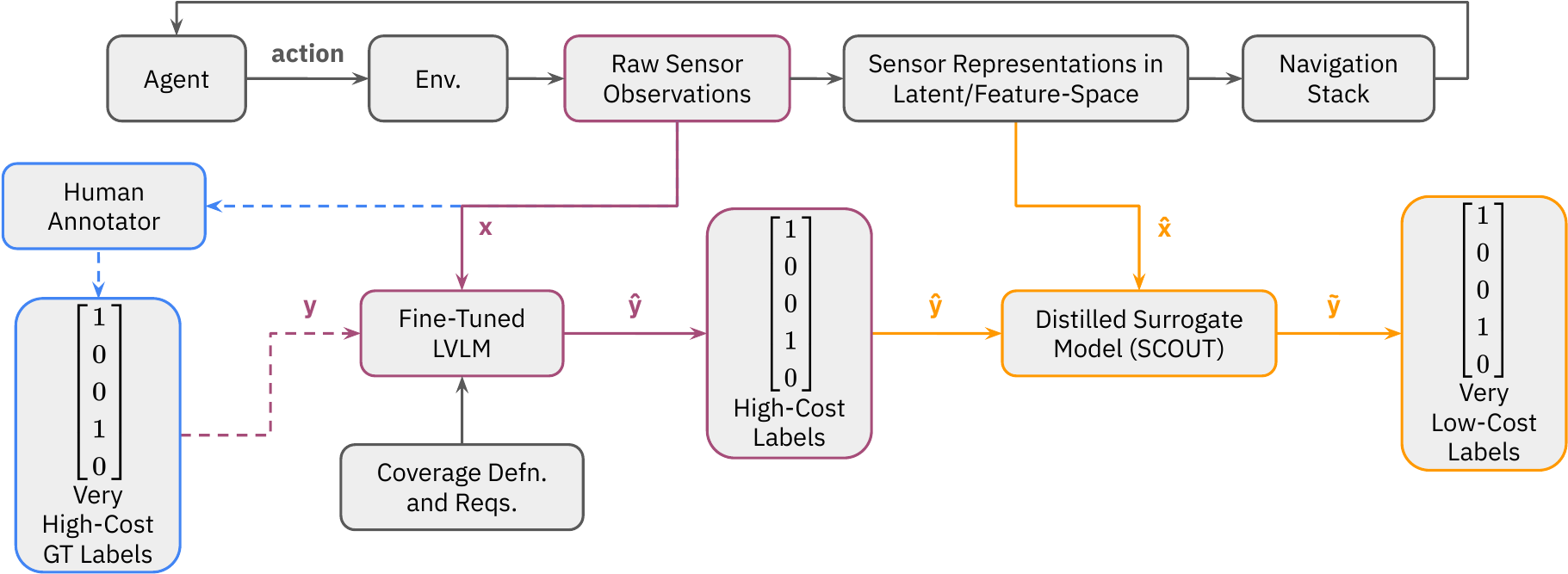}
    \caption{Overview of the scenario coverage pipeline. The distilled surrogate model, SCOUT, predicts scenario coverage labels using precomputed sensor latent representations, which are inherently consumed by the agent's navigation stack.
    Due to the high costs incurred, only a small subset of data is annotated by humans to obtain ground-truth labels.
    To scale the labeling process, an LVLM is fine-tuned and later used to generate labels for a larger dataset, augmenting the training data.
    SCOUT, trained as a distilled surrogate model, learns to replicate the LVLM’s labeling process, thereby enabling lightweight and scalable coverage estimation.}
    \label{fig:scout_pipeline}
\end{figure*}

To this end, we introduce {SCOUT} ({Scenario Coverage Oversight and Understanding Tool}), a surrogate model that efficiently predicts scenario coverage labels using the latent feature representations already computed in an agent's perception stack. 
As illustrated in~\cref{fig:scout_pipeline}, we first fine‑tune an LVLM on a small, human‑labeled subset to generate reliable coverage labels at a larger scale.
SCOUT is then distilled from this LVLM, learning to reproduce coverage labels directly from the agent’s latent sensor features.
Once trained, SCOUT eliminates the need for both human-annotated and LVLM-generated labels, enabling low-cost, high-speed scenario coverage estimation. By leveraging precomputed sensor representations, SCOUT ensures minimal computational overhead while maintaining high accuracy. However, as a surrogate model, its accuracy depends on the quality and diversity of the LVLM-generated labels it is trained on, which may introduce biases if not properly managed. Despite this, SCOUT represents a significant step toward scalable and efficient scenario coverage oversight, making it a practical solution for large-scale autonomous system evaluation.

%% file: Sections/Ch2_RelatedWork.tex
\section{Related Work}

Existing work on scenario coverage can be divided into the following four main categories.

\paragraph{Scenario-Based Test Coverage}  
A growing body of work focuses on formalizing and quantifying coverage for autonomous systems. PhysCov~\cite{hildebrandt2023physcov} introduces a physical test coverage metric by combining vehicle dynamics and sensor inputs to estimate the region of influence during test drives. GUARD~\cite{tu2023towards} proposes a scalable, probabilistic approach to partition scenario parameter spaces without discretization, using Gaussian Processes and level set estimation. Similarly, parameter coverage~\cite{laurent2023parameter} has been introduced to ensure that decision-relevant variables in autonomous driving systems are exercised across diverse simulations. These efforts highlight the growing importance of measuring test sufficiency for safety validation, but they often depend on simulation-specific abstractions or explicit environmental modeling, limiting their generalizability across platforms.

\paragraph{Surrogate Models for Coverage}  
Surrogate modeling has emerged as a powerful tool for reducing the computational cost of evaluating expensive environments. Deep surrogate-assisted methods like DSAGE~\cite{bhatt2022deep} train learned predictors of agent behavior for efficient environment generation. Other works leverage Bayesian optimization with adaptive surrogate models~\cite{lei2021bayesian} or employ neural and Gaussian process-based surrogates for optimizing robot swarm behaviors~\cite{stolfi2023optimising}. RI-SHM~\cite{wu2025ranknet} introduces a surrogate model for mixed-variable coverage optimization problems.
Our study, SCOUT, builds on these ideas, but rather than optimizing control or planning directly, it distills LVLM-labeled coverage indicators into a lightweight predictor that consumes precomputed sensor features, bridging perception and testing in a scalable way.

\paragraph{Classification Based Coverage}  
Several works have tackled the structure and classification of scenario spaces. Tree-based scenario classifications~\cite{schallau2023tree, woodlief2024s3c} introduce a logic-based approach to categorize scenarios over time, enabling systematic analysis of scenario coverage. In parallel, coverage strategies in real-world environments have been studied through multi-agent systems~\cite{patel2020multi}, optimizing urban surveillance using UAVs under sensing and motion constraints. Autonomous monitoring approaches have also been deployed on embedded platforms for specialized detection tasks such as reckless driving~\cite{heo2020autonomous}. While these approaches advance the field of scenario understanding and deployment practicality, SCOUT focuses specifically on labeling coverage itself from latent representations, offering a complementary capability that can plug into broader testing, classification, or deployment pipelines.

\paragraph{Language Model Driven Coverage}  
Recent work has explored the use of Large Language Models (LLMs) for coverage estimation and anomaly detection. LogGPT~\cite{qi2023loggpt} uses an LLM to classify structured system logs via prompt-based reasoning. CSAM \cite{tian2024supervised} integrates an LLM-inspired attention module into an object detection pipeline to improve anomaly detection in cluttered scenes. CoverUp~\cite{pizzorno2024coverup} prompts LLMs to generate high-coverage regression tests by incorporating code and coverage gaps. These efforts demonstrate the growing role of language models in structured reasoning and coverage analysis. SCOUT builds on this direction, distilling LVLM outputs into a lightweight surrogate for scalable, real-time coverage estimation.

%% file: Sections/Ch3_Coverage.tex
\section{Scenario Coverage in Real-World Driving}

Ensuring \textit{scenario coverage} is paramount in evaluating and improving the reliability of autonomous systems deployed in real-world environments. Coverage refers to the degree to which a dataset encompasses all of the possible environmental conditions, behaviors, and hazards in the system’s operational design domain. In safety-critical contexts, insufficient coverage reduces confidence that the system can safely handle encounters with novel or underrepresented situations on public roads or other complex domains.

\subsection{Safety-Critical Relevance}

By training and evaluating a system against diverse operational conditions, including rare and hazardous events,  confidence that the system will safely handle hazardous situations during deployment is improved. However, it is usually impractical to immerse the agent to every possible event.
However, it is usually impractical or unscalable to immerse the agent to \textit{every} unsafe event. Therefore, lightweight and reliable tools are needed to oversee and verify scenario coverage.

\subsection{Definition and Scope of Coverage}

At a high level, coverage can be defined as the \textit{breadth} and \textit{depth} of scenarios that a system has experienced or been validated against. This includes:
\begin{itemize}
    \item High-frequency events, such as straightforward lane-keeping or car-following maneuvers.
    \item Low-frequency but high-severity events, often termed \textit{edge cases}, such as sudden pedestrian intrusions or multi-vehicle collisions on busy highways.
    \item Environmental variations, including weather, lighting, and road types.
    \item Behavioral interactions between vehicles, cyclists, and pedestrians.
\end{itemize}


\subsection{Conflicts in the SHRP2 Naturalistic Driving Study}

An illustrative framework for understanding the kinds of scenarios critical to coverage analysis comes from the second Strategic Highway Research Program (SHRP2)~\cite{hankey2016description,scofield2015researcher}.
SHRP2 identifies \textit{conflicts} as events or situations in which:
\begin{itemize}
    \item A driver or automated system faces an \textit{elevated risk} of collision, road departure, or other hazard.
    \item There is a \textit{noticeable interaction} or potential interaction between traffic participants (e.g., a close following distance or crossing paths at an intersection).
\end{itemize}
During the study, vehicles were instrumented to capture detailed driver behavior and roadway conditions, observing a wide range of conflict scenarios.
SHRP2 introduced a taxonomy of possible conflicts, such as \textit{run-off-road incidents}, \textit{rear-end near-collisions}, \textit{turning across or into traffic}, and \textit{head-on approaches}, each tied to specific driving contexts and geometries.
Through extensive data collection, SHRP2 documented patterns of driver response, near-crash indicators, and collision avoidance behaviors. This categorization has made a formal definition of the scope of scenario coverage for onroad vechicles.

Many conflict types in the SHRP2 taxonomy correspond to rare but critical scenarios.
Consequently, driving datasets often exhibit inherent class imbalance, reflecting true event frequency rather than collection bias. 

\begin{table}[t]
\centering
\caption{Crash typology definitions~\cite{scofield2015researcher} for driving coverage.}
\begin{tabular}{@{}llr@{}}
\toprule
\textbf{Label} & \textbf{Description} & \textbf{Count} \\
\midrule
\multicolumn{3}{l}{\textbf{Group I. Single Driver}} \\[3px]
A & Right roadside departure & 5 \\
B & Left roadside departure & 5 \\
C & Forward impact & 6 \\
\midrule
\multicolumn{3}{l}{\textbf{Group II. Same Trafficway, Same Direction}} \\[3px]
D & Rear end & 17 \\
E & Forward impact & 10 \\
F & Angle/sideswipe & 6 \\
\midrule
\multicolumn{3}{l}{\textbf{Group III. Same Trafficway, Opposite Direction}} \\[3px]
G & Head-on & 4 \\
H & Forward impact & 10 \\
I & Angle/sideswipe & 4 \\
\midrule
\multicolumn{3}{l}{\textbf{Group IV. Change Trafficway, Vehicle Turning}} \\[3px]
J & Turn across path & 8 \\
K & Turn into path & 10 \\
\midrule
\multicolumn{3}{l}{\textbf{Group V. Intersect Paths}} \\[3px]
L & Straight paths & 6 \\
\midrule
\multicolumn{3}{l}{\textbf{Group VI. Misc}} \\[3px]
M & Backing, etc. & 5 \\
\bottomrule
\end{tabular}
\label{tab:crash_typology}
\end{table}

\begin{figure*}[t!]
    \begin{minipage}[b]{0.16\linewidth}
        \centering
        \includegraphics[width=\linewidth]{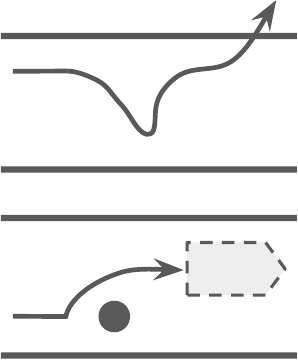}
        \caption{Depictions of example conflicts~\cite{hankey2016description}.}
        \label{fig:sample_shrp2}
    \end{minipage}
    \hfill
    \begin{minipage}[b]{0.77\linewidth}
        \centering
        \includegraphics[width=\linewidth]{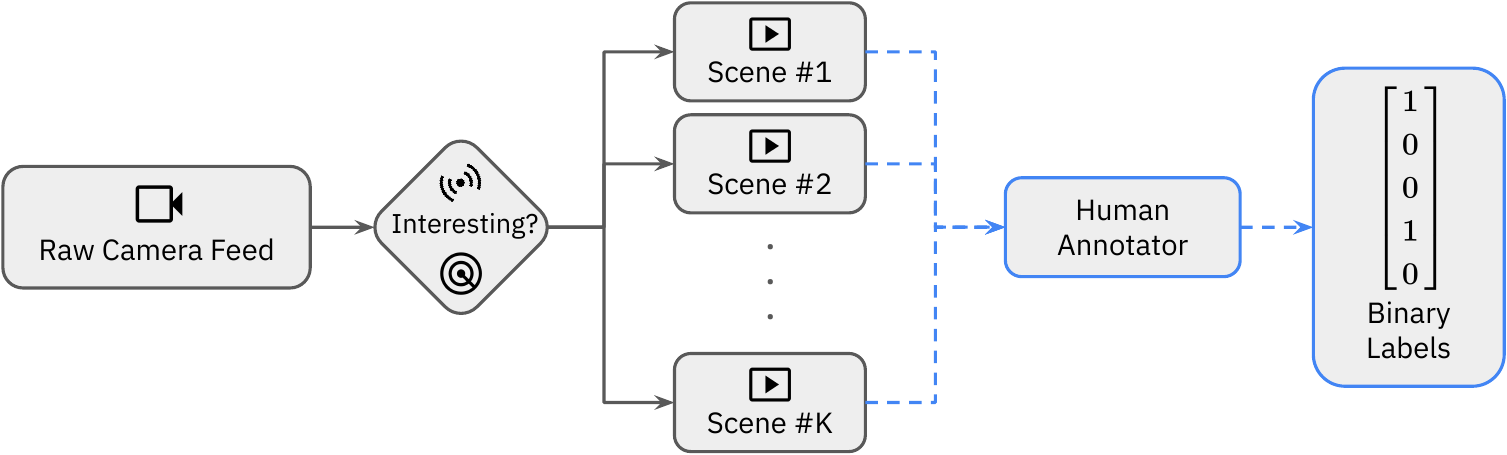}
        \caption{Extraction pipeline of scenes including conflict(s). Raw camera recordings are split the into smaller scenes ($\sim$10~seconds) if they include an \textit{interesting} interaction. A human then annotates them with respect to their context.}
        \label{fig:scene_extraction}
    \end{minipage}%
\end{figure*}

\subsection{Implications for Autonomous Vehicles}

The SHRP2 conflict taxonomy was originally developed to study human driver behavior and roadway safety in naturalistic settings.
The taxonomy, consisting of a total of 95 conflicts, is summarized in~\cref{tab:crash_typology}, where its primary goal is to capture and categorize critical traffic interactions that contribute to crash risk in everyday driving.
Example descriptions from~\cref{tab:crash_typology} are depicted in~\cref{fig:sample_shrp2}. 
The upper example in~\cref{fig:sample_shrp2}, ``\textit{traction loss}'', is one of the 5 descriptions that fall under Group~I.B.
The lower example, \textit{``avoid collision with object''}, is one of the 10 descriptions that fall under Group~II.E.
E.g. The driver in the latter case makes a maneuver to avoid a collision with an onroad object that results in a forward impact with another vehicle in the same trafficway and direction.

In this work, we extend the use case of SHRP2 to the domain of autonomous vehicles (AVs) by leveraging its well-defined categories as a structured protocol for evaluating scenario coverage.
By mapping AV behavior against this taxonomy, we assess whether autonomous systems are exposed to a sufficiently diverse set of realistic and safety-critical driving scenarios, grounded in empirical observations from human driving behavior.

To facilitate this mapping, we implement a human-supervised annotation process depicted in~\cref{fig:scene_extraction}.
Raw forward-facing camera footage is first segmented into shorter clips (i.e. \textit{scenes}) if
an \textit{interesting event} (e.g. includes an interaction that may correspond to one or more SHRP2-defined conflicts) occurs within the raw footage.
These scene instances, typically lasting around 10 seconds each, are detected using other onboard sensor data (e.g. pose, distance to collision, deceleration).
Scenes are then passed to human annotators, who label them with binary indicators reflecting the presence or absence of conflict categories~(\cref{tab:crash_typology}).\linebreak
In this pipeline, expert-designed heuristic-based tools that rely on onboard sensor data may also be leveraged to speed up the manual annotation process.

This annotation process provides high-quality coverage labels that serve as ground truth for our initial training phase of SCOUT.
In doing so, we effectively translate the SHRP2 taxonomy into a practical tool for quantifying AV scenario coverage, allowing us to assess which types of high-risk interactions are underrepresented in a given driving dataset.

%% file: Sections/Ch4_LVLMs.tex
\section{Training Schemas of LLMs and LVLMs}

Large language models (LLMs) have transformed natural language processing by enabling sophisticated understanding, generation, and reasoning. We denote an LLM by $p_\Phi$, which processes a tokenized input sequence 
$ {x = \bigl(x^{(1)}, x^{(2)}, \dots, x^{(n)}\bigr)} $
and models its probability via the factorization
\begin{align}
\label{eq:llm}
p_\Phi(x) 
& = 
p_\Phi\bigl(x^{(1)}, x^{(2)}, \dots, x^{(n)}\bigr) \notag \\
& = \prod_{i=1}^n 
p_\Phi\!\bigl(x^{(i)} \,\bigm|\,
x^{(1)}, \dots, x^{(i-1)}\bigr)
\end{align}
to capture the likelihood of each token $x^{(i)}$ given all preceding tokens in the sequence.

Language models can be extended to include visual embeddings as an additional input modality, giving rise to large vision-language models (LVLMs). In an LVLM, the sequence $x$ may contain tokens from multiple modalities or can include feature tokens extracted from visual inputs (e.g., images). Although the overall modeling framework in~\cref{eq:llm} remains the same, the LVLM incorporates auxiliary components to encode and fuse these multimodal features.

Formally, an LVLM comprises modality encoders $E^{(k)}$, a multimodal fusion module $F$, a language model $f$, and an output projector $P$. Given encoded representations from each modality, the fused embedding is passed through the language model to produce the next-token distribution. At each step $i$, we can write
\begin{align}
\label{eq:lvlm}
p_\Phi(x) 
& = 
p_\Phi\bigl(x^{(1)}, \dots, x^{(n)}\bigr) \notag \\
& = 
P\,\Bigl\{\,
f\,\Bigl[
F\!\bigl(
E^{(1)}(x^{(1)}), \,\dots,\,
E^{(n)}(x^{(n)})
\bigr)
\Bigr]
\Bigr\}
\end{align}
where $x^{(k)}$ denotes the input features (e.g., image embeddings, text tokens) from modality $k$. The fusion module $F(\cdot)$ integrates these modality-specific representations into a unified embedding, $f(\cdot)$ refines it in the language modeling space, and $P(\cdot)$ maps the resulting latent representation to a probability distribution over the next token.

\subsection{Pre-training}
Pre-training is crucial for enabling LLMs
and LVLMs to learn broad linguistic and multimodal representations from large-scale data. 
For the LLM setting, let $\{x_i\}_{i=1}^{N}$ be a collection of unlabeled training sequences, where each sequence ${x_i = \bigl(x_i^{(1)}, \dots, x_i^{(T_i)}\bigr)}$ has length $T_i$. We denote the model parameters by $\Phi$. The goal of pre-training is to maximize the log-likelihood of each token given its preceding context:
\begin{equation}
\Phi^* = \arg\max_{\Phi} \sum_{i=1}^{N} \sum_{j=1}^{T_i}
\log \,p_{\Phi}\bigl(x_i^{(j)}\,\bigm|\,x_i^{(j-c)},\,\dots,\,x_i^{(j-1)}\bigr)
\end{equation}
where $c$ is the context window size that determines how many previous tokens inform the prediction at each position $j$. This formulation also extends naturally to multimodal sequences in LVLMs, where some tokens $x_i^{(j)}$ can represent visual or other modality-specific embeddings where $E$, $F$, $f$, and $P$ are used as shown in~\cref{eq:lvlm}.

\begin{figure*}[t!]
    \centering
    \includegraphics[width=\linewidth]{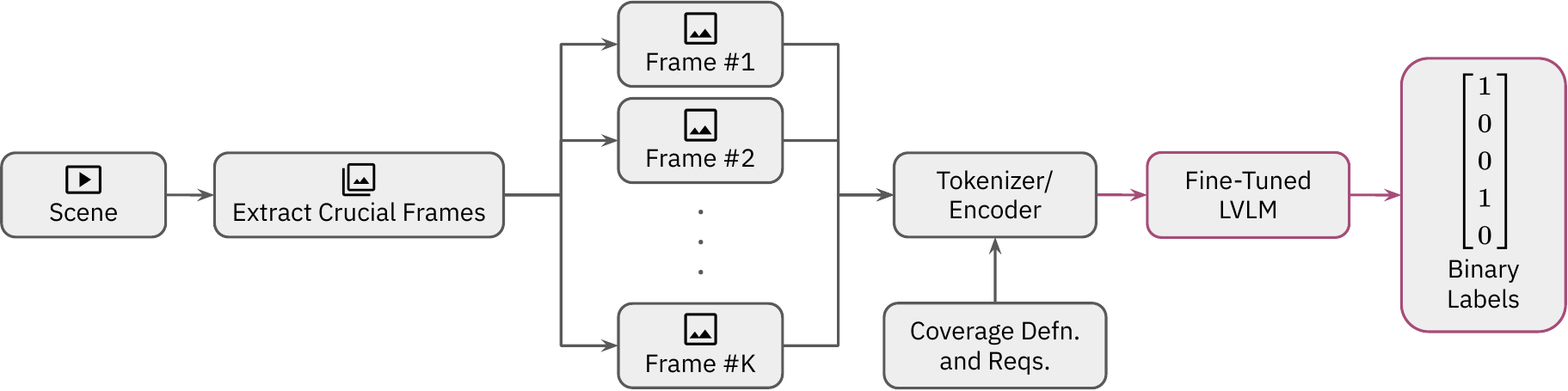}
    \caption{Overview of the scenario coverage label generation pipeline. A driving scene is first processed to extract visually informative frames. Each frame is then encoded, alongside the tokenized/encoded scenario coverage information, and passed through the LVLM. The output is a binary label for each conflict definition, whether they exist within the scene or not.}
    \label{fig:lvlm_frames}
\end{figure*}

\subsection{Fine-tuning}
Fine-tuning adapts a pre-trained model to specific tasks or domains. Previous work explores various techniques, such as prompt tuning \cite{lester2021power}, instruction tuning \cite{zhang2023instruction}, reinforcement learning from human feedback (RLHF) \cite{bai2022training}, and LoRA (low-rank adaptation) \cite{hu2022lora}. In our study, we employ LoRA, which freezes the original model weights and introduces small, trainable matrices $\mathbf{A}_l \in \mathbb{R}^{d \times r}$ and $\mathbf{B}_l \in \mathbb{R}^{r \times d}$ into each Transformer \cite{vaswani2017attention} layer $l$, where $r \ll d$. The adapted weights are computed as
\begin{equation}
\mathbf{W}'_l \;=\; \mathbf{W}_l \;+\; \mathbf{A}_l\,\mathbf{B}_l,
\end{equation}
where $\mathbf{W}_l \in \mathbb{R}^{d \times d}$ are the original (frozen) parameters. LoRA thereby focuses the training on a small subset of parameters $\mathbf{A}_l$ and $\mathbf{B}_l$, which is especially beneficial when fine-tuning large models on specialized tasks.

%% file: Sections/Ch5_Method.tex
\section{Scenario Coverage Oversight and Understanding Tool}

\Cref{fig:scout_pipeline} illustrates the overall pipeline to determine scenario coverage. 
The process begins with an autonomous agent interacting with its environment through an action and collecting raw sensor observations (LiDAR, radar, camera, etc.).
These observations are then transformed into a latent feature representation, which is a standard component of modern navigation stacks. Rather than directly processing raw data, Scenario Coverage Oversight and Understanding Tool (SCOUT) leverages these precomputed representations, ensuring computational efficiency. Training SCOUT is a two-step process.
\begin{enumerate}
    \item Fine-tuning an LVLM to generate a larger training dataset from a small amount of human-labeled (or heuristic based automated labels) ground truth data.
    \item Using the upscaled dataset, train a distilled surrogate model to determine coverage properties for a different input modality, sensor latent representations, rather than their expensive raw versions.
\end{enumerate}

\subsection{Fine-Tuning our LVLM using Human Annotations}

Our primary objective is to train SCOUT, the distilled surrogate model. However, due to the small size of human annotated data available, we first need to augment high-quality but low-cost data.
To do so, we use a pre-trained large vision-language model (LVLM) to that takes in the same modality $x$ as human annotators do, and automatically annotates scenes in a similar fashion. 
Human labeled data $y$ is then used to fine-tune the LVLM to improve its prediction accuracy.
It is important to note that using the LVLM to generate new labels on unseen scenes is still an expensive operation. However, it is cheaper compared to manual hand labels, and thus, is a reasonable alternative to enlarge the data for downstream training.

\Cref{fig:lvlm_frames} outlines the pipeline used to generate scenario coverage labels using an LVLM. The pipeline begins by consuming a scene that was clipped earlier, as shown in~\cref{fig:scene_extraction}.
From this scene, we extract a small number of \textit{crucial frames} (see~\cite{xiao2024method,yasmin2023key}) which are key snapshots that visually capture the main essence of the interactions or conflicts present in the scene.
Each of these extracted frames was then passed through an encoder, which transforms the visual inputs into a format suitable for downstream processing by the LVLM. 

Simultaneously, coverage definitions and requirements, as described in~\cref{tab:crash_typology}, are tokenized and encoded to be passed as a prompt to the LVLM. 
We prompt the model to return a \texttt{Yes} or a \texttt{No} for each conflict definition, which we convert to a binary vector labels $\hat{y}$ afterwards, one for each coverage category predefined in~\cref{tab:crash_typology}.
These labels serve as labels for training the lightweight surrogate model (SCOUT) downstream.

To fine-tune the LVLM, we use LoRA, where the target predictions during training are \texttt{Yes} or \texttt{No}, given the consumed ground truth labels provided by a human annotator (\cref{fig:scene_extraction}).

\subsection{Training a Distilled Surrogate Model}

Once a sufficient volume of scene-level labels has been generated via the fine-tuned LVLM, we proceed to train SCOUT, our distilled surrogate model.
The objective of SCOUT is to replicate the LVLM’s labeling capability while operating on a more efficient input modality: precomputed sensor latent representations.
These latent features, already computed as part of the agent’s navigation stack, are typically available in real-world autonomy pipelines and avoid the need to reprocess raw sensor observations.

We denote the input to SCOUT as~$\hat{x}$, which corresponds to the latent sensor representation for a given driving scene.
SCOUT is trained to predict a binary coverage label vector~$\tilde{y}$ that approximates the high-cost label~$\hat{y}$ produced by the~LVLM.

Training SCOUT requires no additional human annotations or LVLM inference once the distillation process is complete. 
This makes SCOUT an ideal tool for continuous monitoring of scenario coverage during policy evaluation, dataset curation, or simulation-based testing. 
While SCOUT's predictions are inherently bounded by the accuracy of the LVLM it distills, we find that the model maintains high fidelity under realistic conditions, enabling scalable coverage oversight without sacrificing semantic resolution.

%% file: Sections/Ch6_Experiments.tex


\section{Experiments and Results}

\subsection{Dataset, Annotation and Class (Im-)Balance}


We evaluate on a real-world driving corpus of 90{,}000 scenes (5--15 seconds each) collected from an operational fleet of autonomous vehicles. 
A random 10{,}000 scenes were labeled by an expert-designed majorly-automated annotator, producing binary indicators for the 68 conflict types defined by the SHRP2 taxonomy (for Groups II--V in \cref{tab:crash_typology}).\linebreak
We treat these labels as the human-labelled ground truth for this study; the same pipeline would remain unchanged if fully human annotations were substituted. 
The remaining 80{,}000 scenes were labelled by our fine-tuned~LVLM.

In the entire dataset, 45 out of the 68 conflict
categories exhibit a class imbalance worse than 70/30, which is a distribution intentionally preserved to reflect real-world long-tail event scenarios.
Our dataset is split as follows. 
\begin{itemize}
    \item \textbf{Human-labelled split}: 8,000 train / 2,000 test (used \emph{only} for LVLM training).
    \item \textbf{LVLM-labelled split}: 56,000 train / 12,000 val / 12,000 test (used for SCOUT and baseline training).
\end{itemize}

Every scene in the dataset is processed to extract approximately 5 to 8 keyframes (\cref{fig:lvlm_frames}) using Katna~\cite{Katna}.
Latent sensor representations are obtained from the agent’s perception stack.

\subsection{Model and Training Details}

\paragraph{LVLM (Teacher)}
The pre-trained LVLM used for labeling is based on Gemma-3-12B~\cite{team2025gemma}.
The model is fine-tuned through LoRA~\cite{hu2022lora} on SHRP2-aligned coverage definitions~(\cref{tab:crash_typology}) to generate multi-label binary outputs for each scene. 
Unsloth~\cite{Unsloth} is used to reduce memory usage during fine-tuning.

\paragraph{SCOUT (Distilled Surrogate Model)}

SCOUT is a residual~\cite{he2016deep} fully connected neural network (Residual FCNN) with a cross-self-attention mechanism~\cite{vaswani2017attention}. The model processes a sequence of latent embeddings and an attention mask using a multi-head cross-attention layer, followed by mean pooling. The pooled vector passes through three residual blocks, then a projection layer with batch normalization and dropout, before generating multi-label predictions via a sigmoid output.
This design balances semantic expressivity and efficiency, supporting broad generalization with low inference cost.

SCOUT is trained using binary cross-entropy loss over the LVLM-generated labels:
\begin{equation}
\mathcal{L}_{\text{BCE}} = 
- \sum_{i=1}^{G}
\sum_{j=1}^{C}
\left( y_{ij} \log \hat{y}_{ij} + (1 - y_{ij}) \log (1 - \hat{y}_{ij}) \right)
\end{equation}
where $G$ and $C$ are the amounts of coverage categories, and counts in each category, respectively.

\subsection{Evaluating LVLM Agreement with Human Labels}

To validate the quality of our fine-tuned LVLM, we compare its scenario coverage predictions against a manually annotated set of 2,000 scenes labeled by domain experts. This benchmark allows us to assess how well the LVLM aligns with human understanding of the SHRP2 conflict taxonomy.

We compute the following evaluation metrics:
\begin{itemize}
    \item \textbf{Precision, Recall, and F1 Score} for each coverage category.
    \item \textbf{Macro-Averaged F1 Score}, reflecting balanced performance across all categories, regardless of class imbalance.
    \item \textbf{Exact Match Rate}, the proportion of scenes where the LVLM's predicted label vector exactly matches the human annotation.
    \item \textbf{Per-label Agreement Rate}, measuring the percentage of individual category labels that match human annotations across all scenes.
\end{itemize}

\begin{table}[h!]
\centering
\caption{Evaluation of LVLM-predicted scenario coverage labels against human annotations.}
\begin{tabular}{@{}lccr@{}}
\toprule
\textbf{Categories (accumulated)} & \textbf{Precision} & \textbf{Recall} & \textbf{F1 Score} \\
\midrule
Group II. Same TW, Same Dir.             & 0.91 & 0.88 & 0.89 \\
Group III. Same TW, Opp. Dir.    & 0.85 & 0.79 & 0.82 \\
Group IV. Change TW, Veh. Turn.   & 0.87 & 0.75 & 0.80 \\
Group V. Intersect Paths   & 0.83 & 0.86 & 0.84 \\
\midrule
\textbf{Macro Avg.}        & 0.86 & 0.82 & \textbf{0.84} \\
\textbf{Exact Match Rate} & \multicolumn{3}{c}{76.2\%} \\
\textbf{Label Agreement Rate} & \multicolumn{3}{c}{84.5\%} \\
\bottomrule
\end{tabular}
\label{tab:lvml_vs_human}
\end{table}

The quantitative results are summarized in~\cref{tab:lvml_vs_human}, which reports per‑group precision, recall, and F1 scores.
Despite the class imbalance inherent in the training dataset, the LVLM is able to demonstrate a macro averaged F1 score of 0.84 across different groups, indicating strong, balanced performance across the 68 SHRP2 conflict categories.
Moreover, the model achieves an exact match rate and per-label agreement of {76.2\%} and {84.5\%}, respectively, with human annotators.
These findings indicate that the LVLM provides supervision of sufficient fidelity to serve as a teacher model for subsequent distillation. 
We next investigate how closely SCOUT can replicate this performance while operating on sensor‑space embeddings directly.

\begin{table}[b!]
\centering
\caption{SCOUT performance across scenario coverage categories.}
\begin{tabular}{@{}lccr@{}}
\toprule
\textbf{Category (accumulated)} & \textbf{Precision} & \textbf{Recall} & \textbf{F1 Score} \\
\midrule
Group II. Same TW, Same Dir.                 & 0.89 & 0.85 & 0.87 \\
Group III. Same TW, Opp. Dir.    & 0.81 & 0.76 & 0.78 \\
Group IV. Change TW, Veh. Turn.   & 0.79 & 0.72 & 0.75 \\
Group V. Intersect Paths   & 0.80 & 0.84 & 0.82 \\
\midrule
\textbf{Macro Avg.}    & 0.82 & 0.79 & \textbf{0.80} \\
\bottomrule
\end{tabular}
\label{tab:results_main}
\end{table}

\subsection{SCOUT Performance}

SCOUT is trained using the 80,000 scenes labeled by the fine-tuned LVLM.
Table~\ref{tab:results_main} shows SCOUT's performance across the same conflict categories, for the same 2,000 scenes the LVLM is tested on in~\cref{tab:lvml_vs_human}. 
SCOUT is able to achieve a macro averaged F1 score of 0.80, only a 0.04 drop from the fine-tuned LVLM's performance.
Despite being trained on a different modality of input and machine-annotated labels, the distillation process successfully transfers the LVLM’s nuanced coverage reasoning into a much smaller surrogate model.

\subsection{Ablation Study}
\label{subsec:ablation}

We conduct an ablation study to test the impact of different components of SCOUT, by removing one component at a time, and reporting the new macro averaged F1 scores.
We also benchmark against an $\ell_2$-regularized logistic regression model (LogReg) on the same latent features.
These results are shown in Table~\ref{tab:ablation}.  

\begin{table}[htbp]
    \centering
    \caption{Impact of design choices on SCOUT.}
    \begin{tabular}{@{}lcr@{}}
        \toprule
        \textbf{Variant} & \textbf{Macro Avg. F1} & $\Delta$ \textbf{vs.\ Full} \\
        \midrule
        LogReg        & 0.58 & $-\,$0.22 \\
        10\,k training set (instead of 80\,k)        & 0.70 & $-\,$0.10 \\
        No cross‑attention                      & 0.75 & $-\,$0.05 \\
        No dropout                              & 0.77 & $-\,$0.03 \\
        Two residual blocks (instead of three)        & 0.78 & $-\,$0.02 \\
        \textbf{Full SCOUT}                     & \textbf{0.80} &  \\
        \bottomrule
    \end{tabular}
    \label{tab:ablation}
\end{table}

SCOUT’s performance degrades when key components like cross-attention or residual depth are removed, showing their importance for accurate scenario prediction. Using the LVLM to scale up training data proves essential, significantly boosting the surrogate model’s effectiveness.

\subsection{Inference Efficiency}

Table~\ref{tab:inference_cost} compares the inference costs across different methods. Experiments were conducted on an RTX A6000.
SCOUT achieves a massive speedup over both a human annotator as well as a fine-tuned LVLM while having the fraction of memory usage, making real‑time coverage monitoring feasible on‑board the vehicle.

\begin{table}[htbp]
    \centering
    \caption{Inference cost comparisons.}
    \begin{tabular}{@{}lcr@{}}
        \toprule
        \textbf{Model} & \textbf{Avg. Time} & \textbf{VRAM Usage} \\
        \midrule
        Human Annotator & 10–15 min & -- \\
        Fine-Tuned LVLM (Gemma-3-12B)     & 69.4 s & 42.7 GB \\
        \textbf{SCOUT (Distilled Surrogate)}  & \textbf{7.3 s} & \textbf{1.6 GB} \\
        LogReg          & 2.4 s  & 0.4 GB \\
        \bottomrule
    \end{tabular}
    \label{tab:inference_cost}
\end{table}

\subsection{Qualitative Example}

\begin{figure*}[t!]
  \centering
  \includegraphics[width=0.327\textwidth, trim=0 0 0 0, clip]{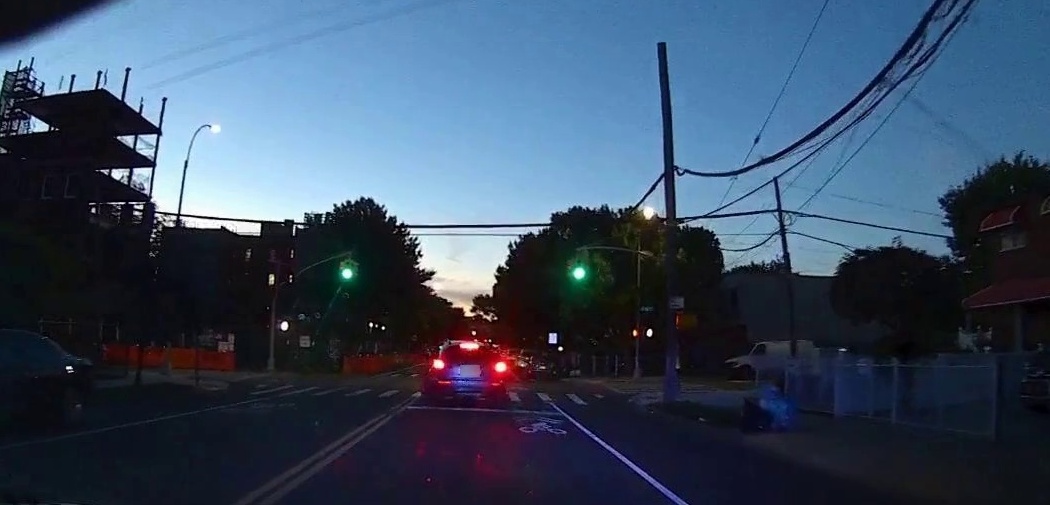}%
  \includegraphics[width=0.327\textwidth, trim=0 0 0 0, clip]{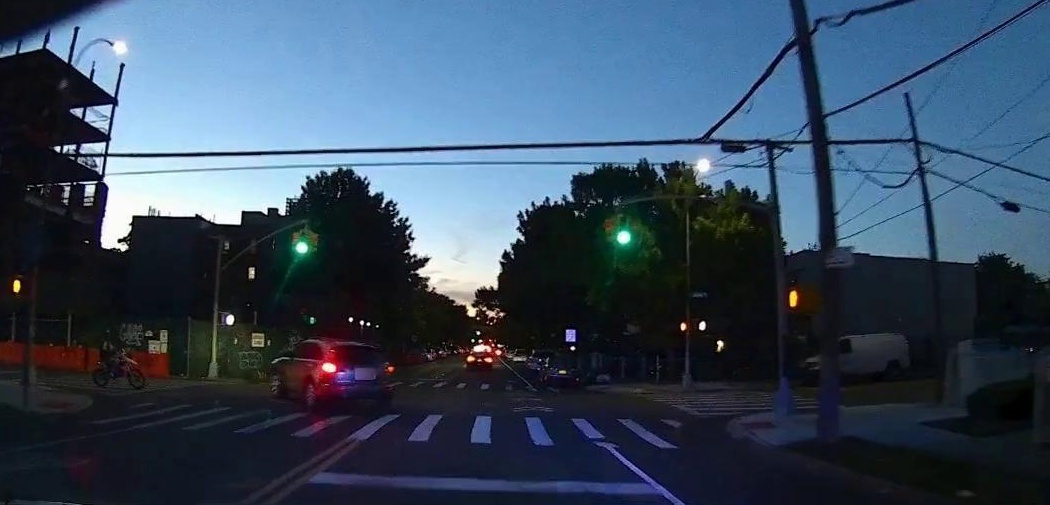}%
  \includegraphics[width=0.327\textwidth, trim=0 0 0 0, clip]{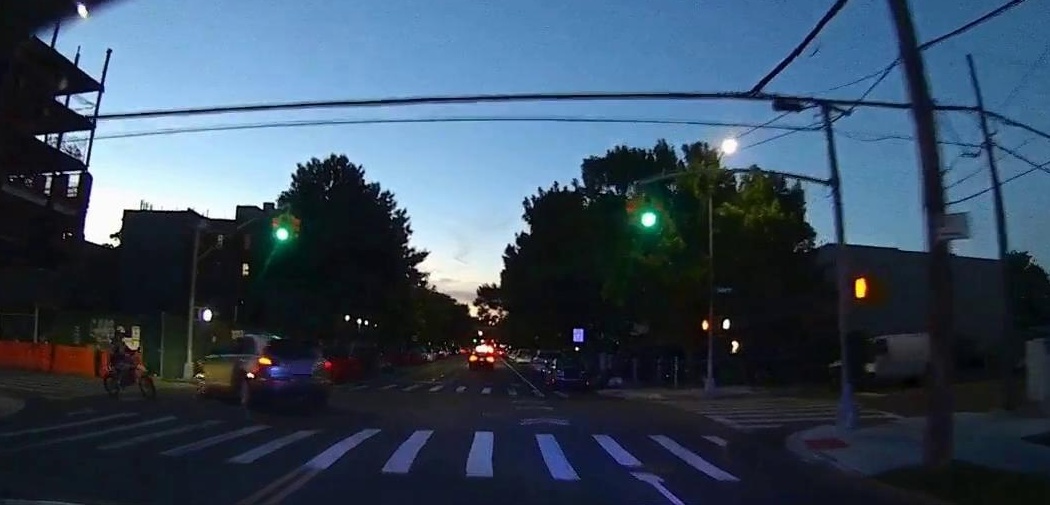}
  \includegraphics[width=0.327\textwidth, trim=0 0 0 0, clip]{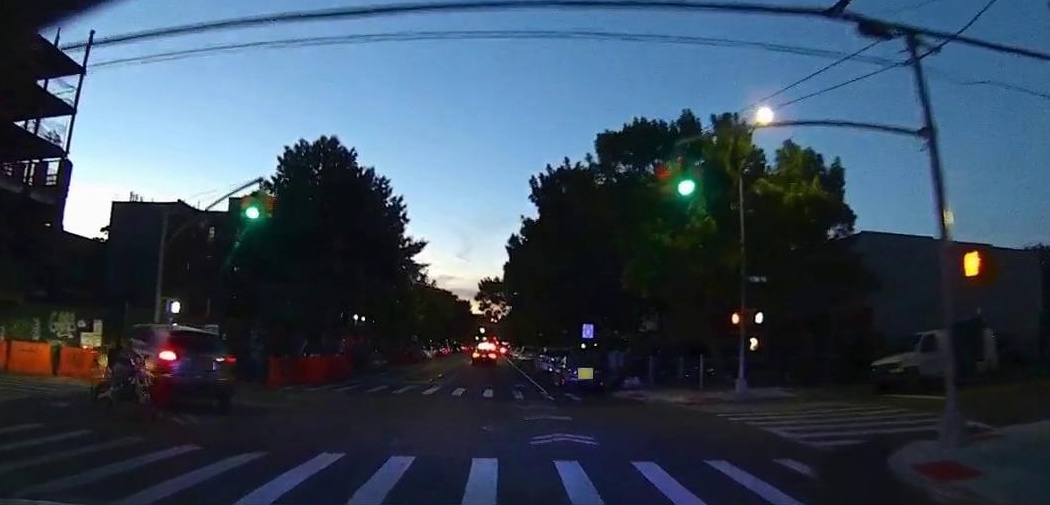}%
  \includegraphics[width=0.327\textwidth, trim=0 0 0 0, clip]{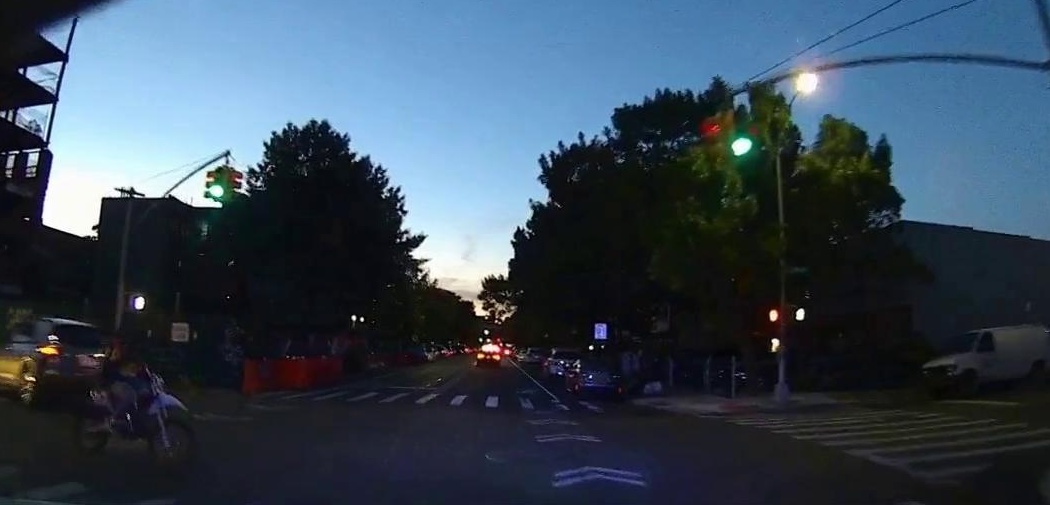}%
  \includegraphics[width=0.327\textwidth, trim=0 0 0 0, clip]{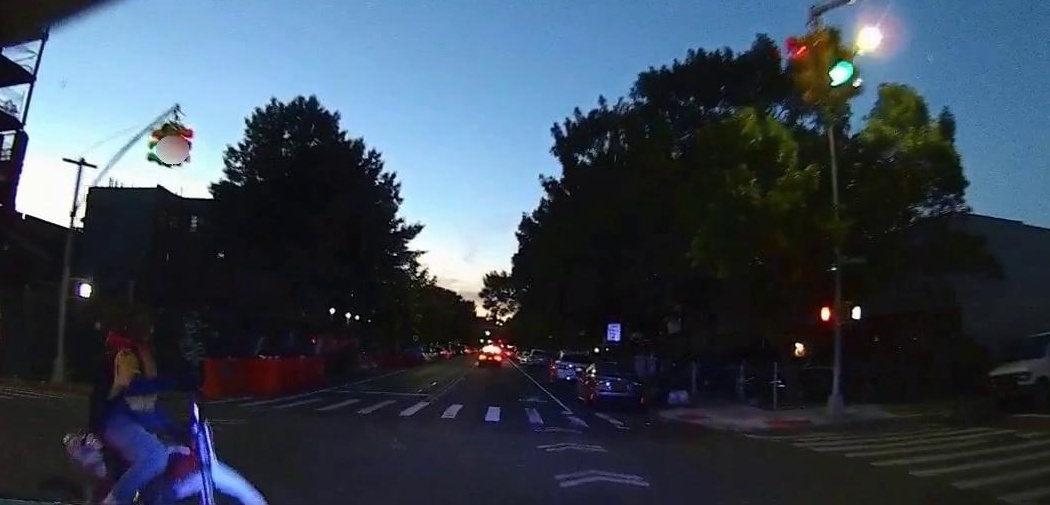}%
  \caption{A driving scene depicting an interaction between a motorcyclist and the ego vehicle, captured from the dashcam perspective. The white motorcycle runs a red light, cutting across the intersection and triggering a high-risk encounter. Frames progress from left to right, starting at the top left and ending at the bottom right.}
  \label{fig:example}
\end{figure*}

\Cref{fig:example} 
shows a scene that was classified by the model as having a safety-critical conflict.
In the scene, a motorbike recklessly enters the four-way intersection the ego vehicle is attempting to cross.
The motorbike illegally crosses at a red light, causing a near-collision with the ego vehicle.
Both vehicles are then able to come to a stop on time before an accident. 
We note that the training and testing data collected by the autonomous fleet always had a safety driver behind the wheel.
For this interaction we correctly identify that
the scene contains a conflict that belongs to a category in Group~V in~\cref{tab:crash_typology}.

%% file: Sections/Ch7_Conclusion.tex
\section{Conclusion and Future Work}

We presented SCOUT, a lightweight surrogate model for estimating scenario coverage in autonomous vehicles using latent features already computed by the agent’s navigation stack. By distilling labels from a fine-tuned LVLM trained on SHRP2-aligned human annotations, SCOUT enables scalable and efficient coverage estimation without relying on costly inference or human labeling.



Experiments across 90,000 real-world driving scenes demonstrate that \textbf{(i)} distillation preserves most of the LVLM’s predictive power, \textbf{(ii)} SCOUT significantly outperforms classical surrogates, and \textbf{(iii)} improvements are statistically robust despite major class imbalance.  
Future work will incorporate temporal localization and semi-supervised self-training.